\documentclass[runningheads]{llncs}
\usepackage{graphicx}

\usepackage{times}
\usepackage{amsmath,bm}
\usepackage{xfrac}
\usepackage{algorithmic}
\usepackage{amsfonts}
\usepackage[inoutnumbered,linesnumbered,ruled,vlined]{algorithm2e}
\usepackage{color}
\usepackage{array}
\usepackage{multirow}
\usepackage{multicol}
\usepackage{tablefootnote}
\usepackage{url}
\usepackage{enumerate}
\newcommand{\camera}[1]{{\color{black}#1}}

\begin{document}
%
\title{Canonicalizing Knowledge Base Literals}
%
%
\author{
Jiaoyan Chen\inst{1} \and
Ernesto Jim\'enez-Ruiz\inst{2,3} \and Ian Horrocks\inst{1,2}
}
%
%
\institute{
Department of Computer Science, University of Oxford, United Kingdom \and
The Alan Turing Institute, United Kingdom \and
Department of Informatics, University of Oslo, Norway
}
\maketitle              
\begin{abstract}
Ontology-based knowledge bases (KBs) like DBpedia are very valuable resources, but their usefulness and usability is limited by various quality issues.
One such issue is the use of string literals instead of semantically typed entities.
In this paper we study the automated \emph{canonicalization} of such literals, i.e.,
replacing the literal with an existing entity from the KB or with a new entity that is typed using classes from the KB. We propose a framework that combines both reasoning and machine learning in order to predict the relevant entities and types, and we evaluate this framework
against state-of-the-art baselines for both semantic typing and entity matching.

\end{abstract}
\section{Introduction}
Ontology-based knowledge bases (KBs) like DBpedia \cite{auer2007dbpedia} are playing an increasingly important role in domains such knowledge management, data analysis and natural language understanding.
Although they are very valuable resources, the usefulness and usability of such KBs is limited by various quality issues \cite{zaveri2016quality,farber2018linked,paulheim2017knowledge}.
One such issue is the use of string literals (both explicitly typed and plain literals) instead of semantically typed entities; for example in the triple $\langle$\textit{River\_Thames}, \textit{passesArea}, \textit{``Port Meadow, Oxford"}$\rangle$. This weakens the KB as it does not capture the semantics of such literals. If, in contrast, the object of the triple were an entity, then this entity could, e.g., be typed as \textit{Wetland} and \textit{Park}, and its location given as \textit{Oxford}.
This problem is pervasive and hence results in a significant loss of information: according to statistics from Gunaratna et al.\ \cite{gunaratna2016gleaning} in 2016, the DBpedia property \textit{dbp:location} has over 105,000 unique string literals that could be matched with entities.
\camera{Besides DBpedia, such literals can also be found in some other KBs from encyclopedias (e.g., zhishi.me \cite{niu2011zhishi}), in RDF graphs transformed from tabular data (e.g., LinkedGeoData \cite{auer2009linkedgeodata}), in aligned or evolving KBs, etc.
}

One possible remedy for this problem is to apply automated semantic typing and entity matching (AKA \emph{canonicalization}\footnote{Note this is different from canonical mapping of literal values in the RDF standard by W3C.}) to such literals. 
To the best of our knowledge, 
semantic typing of KB literals has rarely been studied.
Gunaratna et al.\ \cite{gunaratna2016gleaning} used semantic typing in their entity summarization method, first identifying the so called focus term of a phrase via grammatical structure analysis,
and then matching the focus term with both KB types and entities.
Their method is, however, rather simplistic:
it neither utilizes the literal's context, such as the associated property and subject, 
nor captures the contextual meaning of the relevant words.
What has been widely studied is the semantic annotation of KB entities \cite{gangemi2012automatic,paulheim2013type,sleeman2015entity} and of noun phrases outside the KB (e.g., from web tables) \cite{luo2018cross,efthymiou2017matching,chen2019colnet};
in such cases, however, the context is very different, and
entity typing can, for example, exploit structured information
such as the entity's linked Wikipedia page \cite{gangemi2012automatic}
and the domain and range of properties that the entity is associated with \cite{paulheim2013type}.

With the development of deep learning,
semantic embedding and feature learning
have been widely adopted for exploring different kinds of contextual semantics in prediction, with
Recurrent Neural Network (RNN) being a state-of-the-art method for dealing with structured data and text.
One well known example is \textit{word2vec} --- an RNN language model which can represent words in a vector space that retains their meaning \cite{mikolov2013efficient}.
Another example is a recent study 
by Kartsaklis et al.\ \cite{kartsaklis2018mapping},
which maps text to KB entities with a Long-short Term Memory RNN for textual feature learning.
These methods offer the potential for developing accurate prediction-based methods for KB literal typing and entity matching where the contextual semantics is fully exploited.

In this study, we investigate KB literal canonicalization using a combination of RNN-based learning and semantic technologies. 
We first predict the semantic types of a literal by:
\textit{(i)} identifying candidate classes via lexical entity matching and KB queries;
\textit{(ii)} automatically generating positive and negative examples via KB sampling, with external semantics (e.g., from other KBs) injected for improved quality;
\textit{(iii)} training classifiers using relevant subject-predicate-literal triples embedded in an attentive bidirectional RNN (AttBiRNN);
and \textit{(iv)} using the trained classifiers and KB class hierarchy to predict candidate types.
The novelty of our framework lies in its 
knowledge-based learning; this includes automatic candidate class extraction and sampling from the KB,
triple embedding with different importance degrees suggesting different semantics, and
using the predicted types to identify a potential canonical entity from the KB.
We have evaluated our framework using a synthetic literal set (S-Lite) and a real literal set (R-Lite) from DBpedia \cite{auer2007dbpedia}. The results are very promising, with
significant improvements over several baselines, including the existing state-of-the-art.


\section{Method}\label{sec:method}

\subsection{Problem Statement}

In this study we consider a knowledge base (KB) that includes both
ontological axioms that induce (at least) a hierarchy of semantic types
(i.e., classes), and assertions that describe concrete entities (individuals).
Each such assertion is assumed to be in the form of an RDF triple $\langle s,p,o \rangle$,
where $s$ is an entity, $p$ is a property and $o$ can be either an entity or
a literal (i.e., a typed or untyped data value such as a string or integer).

We focus on triples of the form $\langle s,p,l \rangle$, where $l$ is a string literal; such literals can be identified by regular expressions, as in \cite{gunaratna2016gleaning}, or by
data type inference as in \cite{dongo2017semantic}.
Our aim is to cononicalize $l$ by first identifying the \emph{type} of $l$, i.e., a set of classes $\mathcal{C}_l$ that an entity corresponding to $l$ should be an instance of, and then determining if such an entity already exists in the KB. 
The first subtask is modeled as a machine learning classification problem where a real value score in $\left[0,1\right]$ is assigned to each class $c$ occurring in the KB, and $\mathcal{C}_l$ is the set of classes determined by the assigned score with strategies e.g., adopting a class if its score exceeds some threshold.
The second subtask is modeled as an entity lookup problem constrained by $\mathcal{C}_l$.

It is important to note that:
\begin{enumerate}[(i)]
    \item When we talk about a literal $l$ we mean the occurrence of $l$ in a triple $\langle s,p,l \rangle$. Lexically equivalent literals might be treated very differently depending on their triple contexts.
    \item If the KB is an OWL DL ontology, then the set of \emph{object properties} (which connect two entities) and \emph{data properties} (which connect an entity to a literal) should be disjoint. In practice, however, KBs such as DBpedia often don't respect this constraint. In any case, we avoid the issue by simply computing the relevant typing and canonicalization information, and leaving it up to applications as to how they want to exploit it.
    \item We assume that no manual annotations or external labels are given --- the classifier is automatically trained using the KB.
\end{enumerate}

\subsection{Technical Framework}
\begin{figure}[h]
\centering
\includegraphics[scale=0.45]{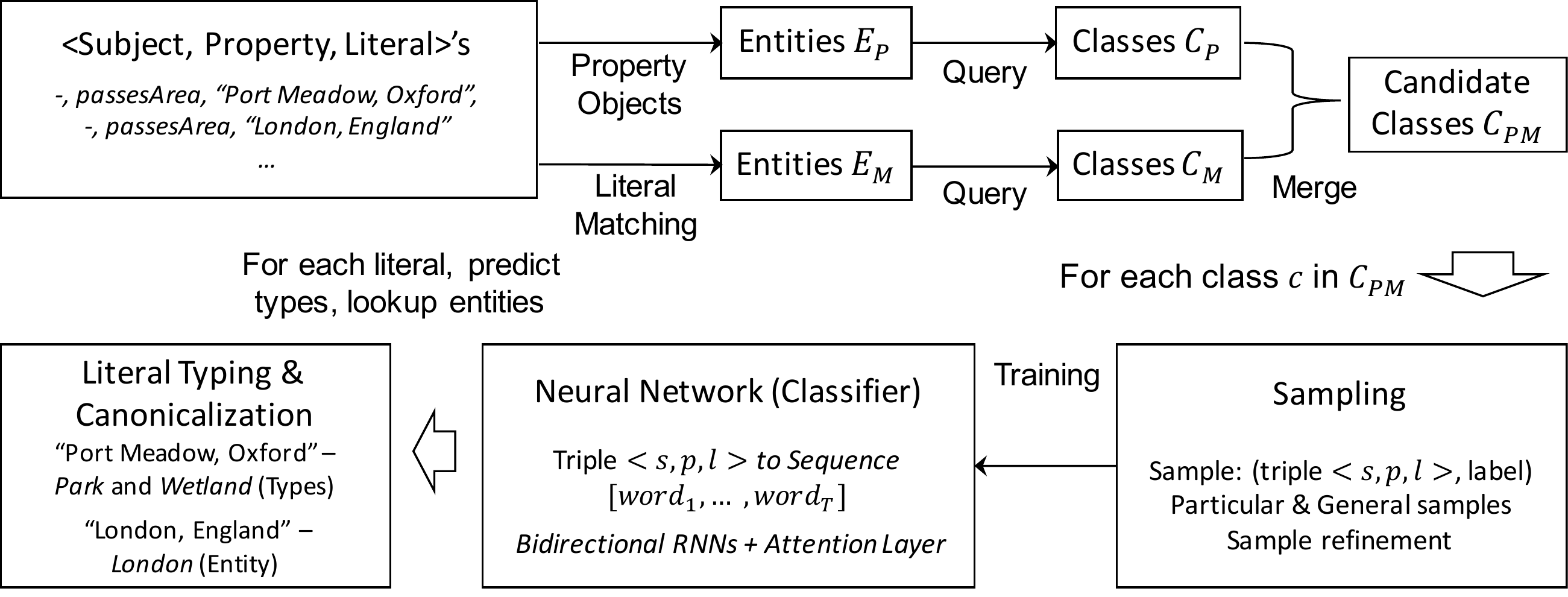}
\caption{\footnotesize
The technical framework for KB literal canonicalization.
}
\label{fig:framework}
\end{figure}

The technical framework for the classification problem is shown in Fig.~\ref{fig:framework}.
It involves three main steps:  
\textit{(i)} candidate class extraction; 
\textit{(ii)} model training and prediction; 
and \textit{(iii)} literal typing and canonicalization.

\subsubsection{Candidate class extraction} 
Popular KBs like DBpedia often contain a large number of classes.
For efficiency reasons, and to reduce noise in the learning process, we first
identify a subset of candidate classes. This selection should be rather inclusive
so as to maximize potential recall. In order to achieve this we pool the candidate
classes for all literals occurring in triples with a given property; i.e., to compute the
candidate classes for a literal $\l$ occurring in a triple $\langle s,p,l \rangle$, 
we consider all triples that use property $p$.
Note that, as discussed above, in practice such triples may include both literals and entities as their objects.
We thus use two techniques for identifying candidate classes from the given set of triples.
In the case where the object of the triple is an entity, the candidates are just the set of classes that this entity is an instance of. In practice we identify the candidates for the set of all such entities, which we denote $E_P$, via a SPARQL query to the KB, with the resulting set of classes being denoted $C_P$.
In the case where the object of the triple is a literal, we first match the literal to entities using
a lexical index which is built based on the entity's name, labels and anchor text (description). 
To maximize recall, the literal, its tokens (words) and its sub-phrases are used to retrieve entities by lexical matching;
this technique is particularly effective when the literal is a long phrase.
As in the first case, we identify all relevant entities, which we denote $E_M$,
and then retrieve the relevant classes $C_M$ using a SPARQL query.
The candidate class set is simply the union of $C_P$ and $C_M$, denoted as $C_{PM}$. 

\subsubsection{Model training and prediction.} 
We adopt the strategy of training one binary classifier for each candidate class, instead of multi-class classification, so as to facilitate dealing with the class hierarchy \cite{silla2011survey}. 
The classifier architecture includes an input layer with word embedding, an encoding layer with bidirectional RNNs, an attention layer and a fully connected (FC) layer for modeling the contextual semantics of the literal.
To train a classifier, both positive and negative entities (samples), including those from $E_M$ (particular samples) and those outside $E_M$ (general samples) are extracted from the KB, with external KBs and logical constraints being used to improve
sample quality.
The trained classifiers are used to compute a score for each candidate class. 

\subsubsection{Literal Typing and Canonicalization} 
The final stage is to semantically type and, where possible, canonicalise literals. 
For a given literal, two strategies, independent and hierarchical, are used to determine its types (classes), with a score for each type. We then use these types and scores to try to identify an entity in the KB that could reasonably be substituted for the literal.

\subsection{Prediction Model}
\begin{figure}[h]
\centering
\includegraphics[scale=0.44]{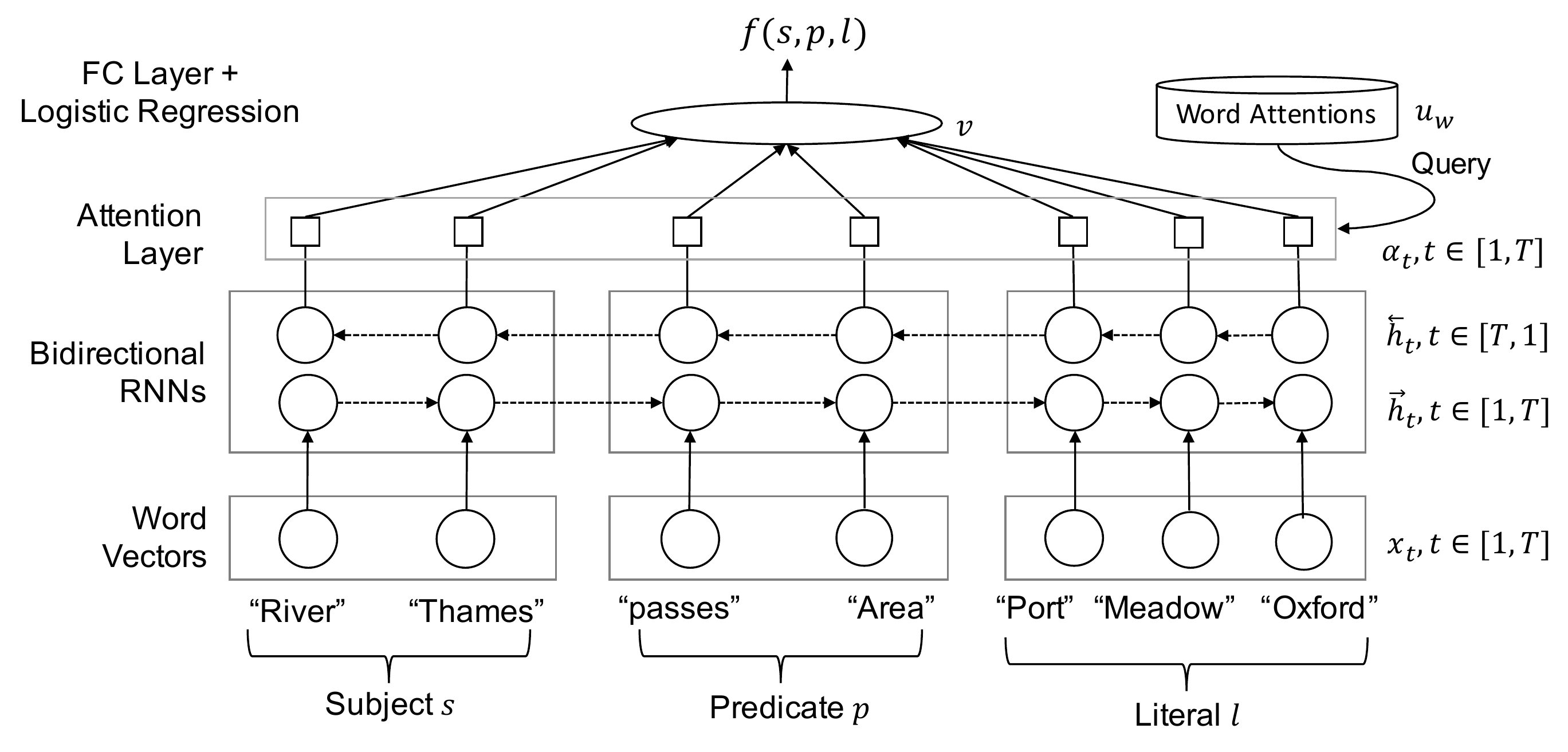}
\caption{\footnotesize
The architecture of the neural network.
}
\label{fig:classifier}
\end{figure}

Given a phrase literal $l$ and its associated RDF triple $\langle s, p, l \rangle$, 
our neural network model aims at utilizing the semantics of $s$, $p$ and $l$ for the classification of $l$.
The architecture is shown in Fig. \ref{fig:classifier}. 
It first separately parses the subject label, the property label and the literal into three word (token) sequences whose lengths, denoted as $T_s$, $T_p$ and $T_l$,
are fixed to the maximum subject, property and literal sequence lengths from the training data by
padding shorter sequences with null words.
We then concatenate the three sequences into a single word sequence ($word_t, t \in \left[1,T\right]$), where $T = T_s + T_p + T_l$.
Each word is then encoded into a vector via word embedding (null is encoded into a zero vector), 
and the word sequence is transformed into a vector sequence ($x_t, t \in \left[1,T\right]$).
Note that this preserves information about the position of words in $s$, $p$ and $l$.

The semantics of forward and backward surrounding words is effective in predicting a word's semantics.
For example, ``Port'' and ``Meadow'' are more likely to indicate a place as they appear after ``Area'' and before ``Oxford''.
To embed such contextual semantics into a feature vector, we stack a layer composed of bidirectional Recurrent Neural Networks (BiRNNs) with Gated Recurrent Unit (GRU) \cite{cho2014learning}.
Within each RNN, 
a reset gate $r_t$ is used to control the contribution of the past word, 
and an update gate $z_t$ is used to balance the contributions of the past words and the new words.
The hidden state (embedding) at position $t$ is computed as
\begin{equation}\label{eq:rnn}
\begin{cases}
h_t = (1-z_t) \odot h_{t-1} + z_t \odot \tilde{h}_t, \\
\tilde{h}_t = 
\tau(W_h x_t + r_t \odot (U_h h_{t-1}) + b_h), \\
z_t = \sigma(W_z x_t + U_z h_{t-1} + b_z), \\
r_t = \sigma(W_r x_t + U_r h_{t-1} + b_r),
\end{cases}
\end{equation}
where $\odot$ denotes the Hadamard product,
$\sigma$ and $\tau$ denote the activation function of \textit{sigmod} and \textit{tanh} respectively, and
$W_h$, $U_h$, $b_h$, $W_z$, $U_z$, $b_z$, $W_r$, $U_r$ and $b_r$ are parameters to learn.
With the two bidirectional RNNs, one forward hidden state and one backward hidden state are calculated for the sequence, denoted as ($\overrightarrow{h_t}
, t \in \left[1,T \right]$) and ($\overleftarrow{h_t}
, t \in \left[T,1 \right]$) respectively.
They are concatenated as the output of the RNN layer: $h_t = \left[\overrightarrow{h_t}, \overleftarrow{h_t}\right], t \in \left[1,T \right]$.

We assume different words are differently informative towards the type of the literal. 
For example, the word ``port'' is more important than the other words in distinguishing the type \textit{Wetland} from other concrete types of \textit{Place}. 
To this end, an attention layer is further stacked.
Given the input from the RNN layer  ($h_t, t \in \left[1,T \right]$), 
the attention layer outputs 
$h_a = \left[\alpha_t h_t \right], t \in \left[1,T \right]$, 
where $\alpha_t$ is the normalized weight of the word at position $t$ and is calculated as 
\begin{equation}\label{eq:att}
\begin{cases}
\alpha_t = \frac{exp(u^T_t u_w)}{\sum_{t \in \left[1,T\right]} exp (u^T_t u_w)} \\
u_t = 
\tau(W_w h_t + b_w),
\end{cases}
\end{equation}
where $u_w$, $W_w$ and $b_w$ are parameters to learn.
Specifically, $u_w$ denotes the general informative degrees of all the words, 
while $\alpha_t$ denotes the attention of the word at position $t$ w.r.t. other words in the sequence.
Note that the attention weights can also be utilized to justify a prediction.
In order to exploit information about the location of a word in the subject, property or literal, we do not calculate the weighted sum of the BiRNN output but 
concatenate the weighted vectors.
The dimension of each RNN hidden state (i.e., $\overleftarrow{h_t}$ and $\overrightarrow{h_t}$), denoted as $d_r$,
and the dimension of each attention layer output (i.e., $\alpha_t h_t$), denoted as $d_a$, are two hyper parameters of the network architecture.

A fully connected (FC) layer and a logistic regression layer are finally stacked for modeling the nonlinear relationship and calculating the output score respectively:
\begin{equation} \label{eq:fc}
   f(s, p, l) = \sigma(W_f h_a + b_f),
\end{equation}
where $W_f$ and $b_f$ are the parameters to learn, $\sigma$ denotes the \textit{sigmod} function, and $f$ denotes the function of the whole network.

\subsection{Sampling and Training}

We first extract both particular samples
and general samples from the KB using SPARQL queries and reasoning;
we then improve sample quality by detecting and repairing wrong and missing entity classifications with the help of external KBs;
and finally we train the classifiers.

\subsubsection{Particular Sample} Particular samples are based on the entities $E_M$ that are lexically matched by the literals.
For each literal candidate class $c$ in $C_M$,
its particular samples are generated by: 
\begin{enumerate}[(i)]
\item Extracting its positive particular entities: $E_M^c = \left\{e | e \in E_M, e \text{ is an instance of } c \right\}$;
\item Generating its positive particular samples as 
\begin{equation}\label{eq:sample}
\mathcal{P}_c^{+} = \cup_{e \in E_M^c} \left\{ \langle s,p,l \rangle | s \in S(p,e), l \in L(e) \right\},
\end{equation}
where $S(p,e)$ denotes the set of entities occurring in the subject position in a triple of the form $\langle s, p, e\rangle$, and $L(e)$ denotes all the labels (text phrases) of the entity $e$;
\item Extracting its negative particular entities $E_M^{\widetilde{c}}$ as those entities in $E_M$ that are instances of some sibling class of $c$ and not instances of $c$;\footnote{We use sibling classes to generate negative examples as, in practice, sibling classes are often disjoint.}
\item Generating its negative particular samples $\mathcal{P}_c^-$ with $E_M^{\widetilde{c}}$
using the same approach as for positive samples.
\end{enumerate}

\subsubsection{General Sample} Given that the literal matched candidate classes $C_M$ are only a part of all the candidate classes $C_{PM}$,
and that the size of particular samples may be too small to train the neural network, 
we additionally generate general samples based on common KB entities.
For each candidate class $c$ in $C_{PM}$, 
all its entities in the KB, denoted as $E^c$, are extracted
and then its positive general samples, denoted as $\mathcal{G}_c^+$,
are generated from $E^c$ using the same approach as for particular samples.
%
Similarly, entities of the sibling classes of $c$, denoted as $E^{\widetilde{c}}$, are extracted, 
and general negative samples, denoted as $\mathcal{G}_c^-$, are generated from $E^{\widetilde{c}}$.
As for negative particular entities, 
we check each entity in $E^{\widetilde{c}}$ and remove those that are not instances of $c$.

Unlike the particular samples, the positive and negative general samples are balanced.
This means that we reduce the size of $\mathcal{G}_c^+$ and $\mathcal{G}_c^-$ to
the minimum of $\#(\mathcal{G}_c^+)$, $\#(\mathcal{G}_c^-)$ and $N_0$, where $\#()$ denotes set cardinality, and $N_0$ is a hyper parameter for sampling.
Size reduction is implemented via random sampling.

\subsubsection{Sample Refinement}
Many KBs are quite noisy, with wrong or missing entity classifications. 
For example, when using the SPARQL endpoint of DBpedia, \textit{dbr:Scotland} is classified as \textit{dbo:MusicalArtist} instead of as \textit{dbo:Country}, while \textit{dbr:Afghan} appears without a type. We have corrected and complemented the sample generation by combining the outputs of more than one KB. For example, the DBpedia endpoint suggestions are compared against  
Wikidata and the DBpedia lookup service. Most DBpedia entities are mapped to Wikidata entities whose types are used to validate and complement the suggested types from the DBpedia endpoint. In addition, the lookup service, although incomplete, typically provides very precise types that can also confirm the validity of the DBpedia endpoint types. The validation is performed by identifying if the types suggested by one KB are compatible with those returned by other KBs, that is, if the relevant types belong to the same branch of the hierarchy (e.g., the DBpedia taxonomy).
With the new entity classifications, the samples are revised accordingly.

\subsubsection{Training}
We train a binary classifier $f^c$ for each class $c$ in $C_{PM}$.
It is first pre-trained with general samples $\mathcal{G}_{c}^+ \cup \mathcal{G}_{c}^-$,
and then fine tuned with particular samples $\mathcal{P}_{c}^+ \cup \mathcal{P}_{c}^-$.
Pre-training deals with the shortage of particular samples, while fine-tuning bridges the gap between common KB entities and the entities associated with the literals, which is also known as \textit{domain adaptation}.
Given that pre-training is the most time consuming step, but is task agnostic, 
classifiers for all the classes in a KB could be pre-trained in advance to accelerate a specific literal canonicalization task.

\subsection{Independent and Hierarchical Typing}

In prediction, 
the binary classifier for class $c$, denoted as $f^c$,
outputs a score $y_l^c$ indicating the probability that a literal $l$ belongs to class $c$: $y_l^c = f^c(l)$, $y_l^c \in \left[0,1\right]$.
With the predicted scores, we adopt two strategies -- \textit{independent} and \textit{hierarchical} to determine the types.
In the independent strategy, 
the relationship between classes is not considered.
A class $c$ is selected as a type of $l$ if its score $y_l^c \ge \theta$, 
where $\theta$ is a threshold hyper parameter in $[0,1]$. 

The hierarchical strategy considers the class hierarchy and the disjointness between sibling classes. 
We first calculate a \textit{hierarchical score} for each class with the predicted scores of itself and its descendents:
\begin{equation}\label{eq:score}
   s_l^c = max\left\{ y_l^{c'} | c' \sqsubseteq c,\text{ } c' \in C_{PM} \right\},
\end{equation}
where $\sqsubseteq$ denotes the subclass relationship between two classes, $C_{PM}$ is the set of candidate classes for $l$, and $max$ denotes the maximum value of a set.
For a candidate class $c'$ in $C_{PM}$, we denote all disjoint candidate classes as $\mathcal{D}(C_{PM}, c')$.
They can be defined as sibling classes of both $c'$ and its ancestors,
or via logical constraints in the KB. 
A class $c$ is selected as a type of $l$ if \textit{(i)} its hierarchical score $s_l^c \ge \theta$, 
and \textit{(ii)} it satisfies the following soft exclusion condition:
\begin{equation}\label{eq:exclusion}
s_l^c - max\left\{s_l^{c'} | c' \in \mathcal{D}(C_{PM}, c) \right\} \ge \kappa,
\end{equation}
where $\kappa$ is a relaxation hyper parameter.
The exclusion of disjoint classes is hard if $\kappa$ is set to $0$, and relaxed if $\kappa$ is set to a negative float with a small absolute value e.g., $-0.1$.

Finally, for a given literal $l$, we return the set of all selected classes
as its types $\mathcal{C}_l$.

\subsection{Canonicalization}
Given a literal $l$, we use $\mathcal{C}_l$ to try to identify an associated entity.
A set of candidate entities are first retrieved using the lexical index that is built on the entity's name, label, anchor text, etc.
Unlike candidate class extraction, here we use the whole text phrase of the literal, 
and rank the candidate entities according to their lexical similarities.
Those entities that are not instances of any classes in $\mathcal{C}_l$ are then filtered out,
and the most similar entity among the remainder is selected as the associated entity for $l$.
If no entities are retrieved, or all the retrieved entities are filtered out,
then the literal could be associated with a new entity whose types are those most specific classes in $\mathcal{C}_l$.
In either case we can improve the quality of our results by checking that the resulting entities would be consistent if added to the KB, and discarding any entity associations that would lead to inconsistency.

\section{Evaluation}\label{sec:evaluation}

\subsection{Experiment Setting}

\subsubsection{Data Sets}
In the experiments, we adopt a real literal set (R-Lite) and a synthetic literal set (S-Lite)\footnote{Data and codes: 
\url{https://github.com/ChenJiaoyan/KG_Curation}
}
, both of which are extracted from DBpedia.
R-Lite is based on the property and literal pairs published by Gunaratna et al. in 2016 \cite{gunaratna2016gleaning}.
We refine the data by \textit{(i)} removing literals that no longer exist in the current version of DBpedia; 
\textit{(ii)} extracting new literals from DBpedia for properties whose existing literals were all removed in step \textit{(i)};
\textit{(iii)} extending each property and literal pair with an associated subject; and
\textit{(iv)} manually adding ground truth types selected from classes defined in the DBpedia Ontology (DBO)\footnote{Classes with the prefix of http://dbpedia.org/ontology/.}.
To fully evaluate the study with more data,
we additionally constructed S-Lite from DBpedia
by repeatedly: \textit{(i)} selecting a DBpedia triple of the form $\langle s,p,e \rangle$, where $e$ is an entity;
\textit{(ii)} replacing $e$ with it's label $l$ to give a triple $\langle s,p,l \rangle$;
\textit{(iii)} eliminating the entity $e$ from DBpedia; and
\textit{(iv)} adding as ground truth types the DBpedia classes of which $e$ is (implicitly) an instance.
More data details are shown in Table~\ref{res:statistics}.
%
\begin{table}[h!]
\scriptsize{
\centering
\renewcommand{\arraystretch}{1.3}
\begin{tabular}[t]{p{1.1cm}<{\centering}|p{1.35cm}<{\centering}|p{1.35cm}<{\centering}|p{3.8cm}<{\centering}|p{3.6cm}<{\centering}}\hline
& Properties \# & Literals \# & Ground Truth Types \# (per Literal) & Characters (Tokens) \# per Literal \\ \hline
S-Lite & $41$ & $1746$ & $256$ ($2.94$) & $16.66$ ($2.40$) \\ \hline
R-Lite & $142$ & $820$ & $123$ ($3.11$) & $19.44$ ($3.25$) \\ \hline
\end{tabular}
\vspace{0.15cm}
\caption{\footnotesize
Statistics of S-Lite and R-Lite.
}\label{res:statistics}
}
\end{table}

\subsubsection{Metrics}
In evaluating the typing performance, Precision, Recall and F1 Score are used.
For a literal $l$, the computed types $\mathcal{C}_l$ are compared with the ground truths $\mathcal{C}_l^{gt}$,
and the following micro metrics are calculated:
$P_l = \sfrac{\# (\mathcal{C}_l \cap \mathcal{C}_l^{gt}) }{\# (\mathcal{C}_l)}$,
$R_l = \sfrac{\# (\mathcal{C}_l \cap \mathcal{C}_l^{gt} )}{\# (\mathcal{C}_l^{gt})}$, and 
${F_1}_l = \sfrac{(2 \times P_l \times R_l)}{(P_l + R_l)}$.
They are then averaged over all the literals as the final Precision, Recall and F1 Score of a literal set.
Although F1 Score measures the overall performance with both Precision and Recall considered,
it depends on the threshold hyper parameter $\theta$ as with Precision and Recall.
Thus we let $\theta$ range from $0$ to $1$ with a step of $0.01$, and calculate the average of all the F1 Scores (AvgF1@all) and top $5$ highest F1 Scores (AvgF1@top$5$).
AvgF1@all measures the overall pattern recognition capability, while AvgF1@top$5$ is relevant in real applications where we often use a validation data set to find a $\theta$ setting that is close to the optimum. 
We also use the highest (top) Precision in evaluating the sample refinement.

In evaluating entity matching performance, Precision is measured
by manually checking whether the identified entity is correct or not.
S-Lite is not used for entity matching evaluation as the corresponding entities for all its literals are assumed to be excluded from the KB.
We are not able to measure recall for entity matching
as we do not have the ground truths; instead, we have evaluated entity
matching with different confidence thresholds and compared the number of correct results.

\subsubsection{Baselines and Settings}
The evaluation includes three aspects. 
We first compare different settings of the typing framework, analyzing the impacts of sample refinement, fine tuning by particular samples, BiRNN and the attention mechanism.
We also compare the independent and hierarchical typing strategies.
We then compare the overall typing performance of our framework with \textit{(i)} Gunaratna et al.\ \cite{gunaratna2016gleaning}, which matches the literal to both classes and entities; 
\textit{(ii)} an entity lookup based method;
and \textit{(iii)} a probabilistic property range estimation method.
Finally, we analyze the performance of entity matching with and without the predicted types.

The DBpedia lookup service, which is based on the Spotlight index \cite{mendes2011dbpedia}, is used for entity lookup (retrieval). 
The DBpedia SPARQL endpoint is used for query answering and reasoning.
The reported results are based on the following settings: the Adam optimizer together with cross-entropy loss are used for network training; 
$d_r$ and $d_a$ are set to $200$ and $50$ respectively;
$N_0$ is set to $1200$;
\textit{word2vec} trained with the latest Wikipedia article dump is adopted for word embedding; 
and ($T_s$, $T_p$, $T_l$) are set to ($12$, $4$, $12$) for S-Lite and ($12$, $4$, $15$) for R-Lite.
The experiments are run on a workstation with Intel(R) Xeon(R) CPU E5-2670 @2.60GHz, 
with programs implemented by Tensorflow.

\subsection{Results on Framework Settings}\label{sec:framework}
We first evaluate the impact of the neural network architecture, fine tuning and different typing strategies, with their typing results on S-Lite shown in Table \ref{res:setting} and Fig.~\ref{res:settings2}.
Our findings are supported by comparable results on R-Lite.
We further evaluate sample refinement, with some statistics of the refinement operations as well as performance improvements shown in Fig.~\ref{res:settings3}.

\subsubsection{Network Architecture and Fine Tuning}
According to Table \ref{res:setting}, we find BiRNN significantly outperforms Multiple Layer Perceptron (MLP), a basic but widely used neural network model, 
while stacking an attention layer (AttBiRNN) further improves AvgF1@all and AvgF1@top$5$, for example by $3.7\%$ and $3.1\%$ respectively with hierarchical typing ($\kappa$ = $-0.1$).
The result is consistent for both pre-trained models and fine tuned models, using both independent and hierarchical typing strategies.
This indicates the effectiveness of our neural network architecture.
Meanwhile, the performance of all the models is significantly improved after they are fine tuned by the particular samples, as expected.
For example, when the independent typing strategy is used,
AvgF1@all and AvgF1@top$5$ of AttBiRNN are improved by $54.1\%$ and $35.2\%$ respectively.
\vspace{-0.3cm}
\begin{table}[h!]
\scriptsize{
\centering
\renewcommand{\arraystretch}{1.3}
\begin{tabular}[t]{p{1.5cm}<{\centering}|p{1.2cm}<{\centering}||p{1.32cm}<{\centering}|c||p{1.32cm}<{\centering}|c||p{1.32cm}<{\centering}|c}
\hline
 \multicolumn{2}{c||}{Framework} & \multicolumn{2}{c||}{Independent} & \multicolumn{2}{c||}{Hierarchical ($\kappa=-0.1$)} & \multicolumn{2}{c}{Hierarchical ($\kappa=0$)}    \\\cline{3-8}
 \multicolumn{2}{c||}{Settings} & AvgF1@all & AvgF1@top$5$ & AvgF1@all & AvgF1@top$5$& AvgF1@all & AvgF1@top$5$ \\ \hline
 \multirow{3}{*}{Pre-training} & MLP &$0.4102$ & $0.4832$ &$0.5060$  &$0.5458$  &$0.5916$  &$0.5923$  \\ 
  & BiRNN &$0.4686$ & $0.5566$ & $0.5295$ & $0.5649$ & $0.5977$ & $0.5985$ \\ 
 & AttBiRNN & $0.4728$ & $0.5590$ & $0.5420$	& $0.5912$ & $0.6049$ & $0.6052$ \\ \cline{1-8}
 \multirow{3}{*}{Fine tuning} & MLP &$0.6506$ &$0.6948$ &$0.6859$ &$0.6989$&$0.6429$ &$0.6626$  \\
  & BiRNN & $0.7008$	& $0.7434$	& $0.7167$ &	$0.7372$	& $0.6697$	& $0.6850$ \\
  & AttBiRNN & $0.7286$	& $0.7557$ &$0.7429$	& $0.7601$	& $0.6918$	& $0.7070$ \\ \cline{1-8}
\end{tabular}
\vspace{0.2cm}
\caption{\footnotesize
Typing performance of our framework on S-Lite under different settings.
}\label{res:setting}
}
\end{table}
\vspace{-0.8cm}

\subsubsection{Independent and Hierarchical Typing} 
\begin{figure}[h]
\centering
\includegraphics[scale=0.35]{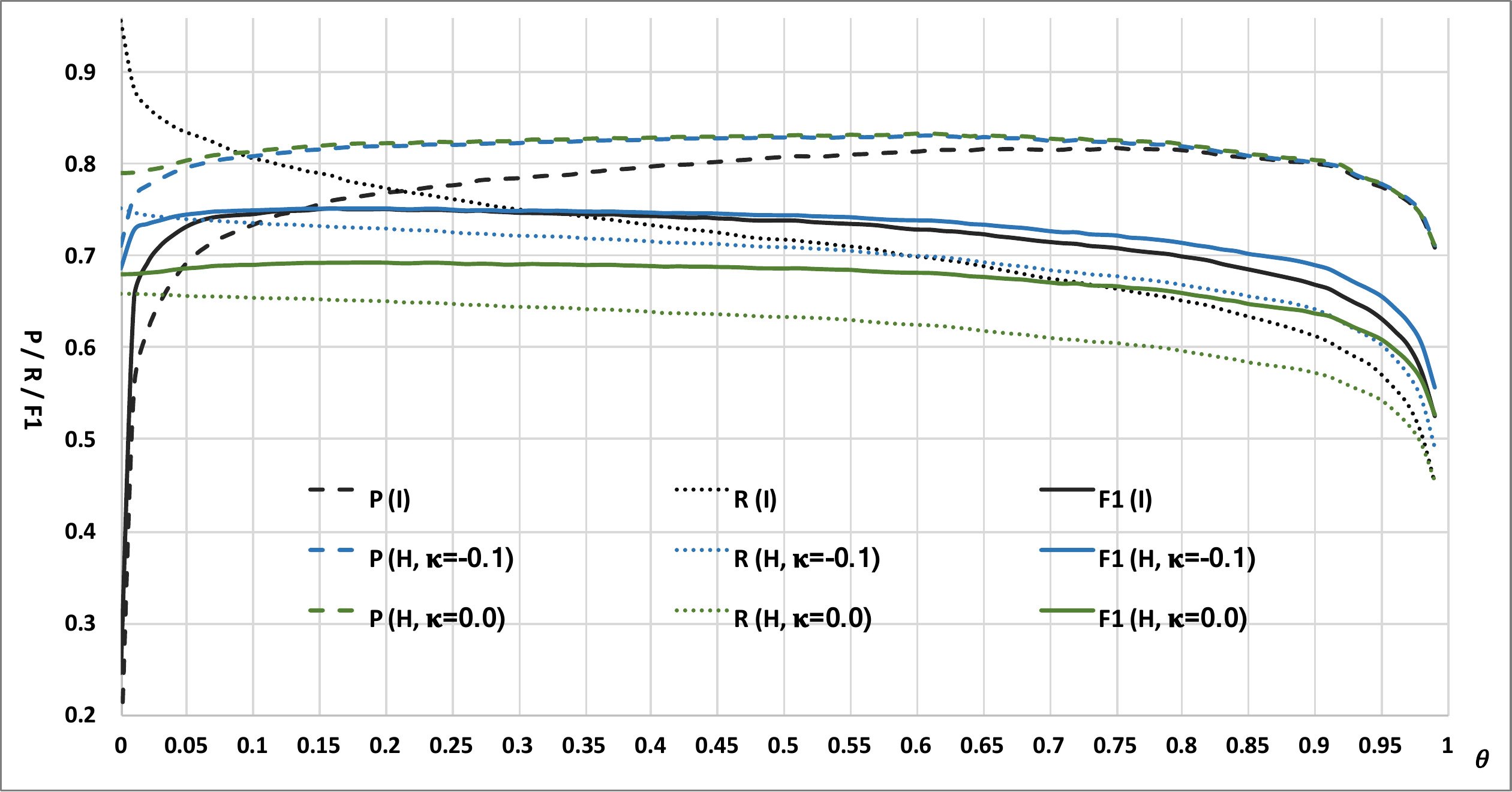}
\caption{\footnotesize
(P)recision, (R)ecall and (F1) Score of independent (I) and hierarchical (H) typing for S-Lite,
with the scores predicted by the fine tuned AttBiRNN.
}
\label{res:settings2}
\end{figure}

The impact of independent and hierarchical typing strategies is more complex.
As shown in Table \ref{res:setting}, when the classifier is weak (e.g., pre-trained BiRNN), hierarchical typing with both hard exclusion ($\kappa$ = $0$) and relaxed exclusion ($\kappa$ = $-0.1$) has higher AvgF1@all and AvgF1@top$5$ than independent typing.
However, when a strong classifier (e.g., fine tuned AttBiRNN) is used, AvgF1@all and AvgF1@top$5$ of hierarchical typing with relaxed exclusion are close to independent typing, 
while hierarchical typing with hard exclusion has worse performance.
We further analyze Precision, Recall and F1 Score of both typing strategies under varying threshold ($\theta$) values, as shown in Fig.~\ref{res:settings2}.
In comparison with independent typing, hierarchical typing achieves \textit{(i)} more stable Precision, Recall and F1 Score curves; and \textit{(ii)} significantly higher Precision, especially when $\theta$ is small.
Meanwhile, as with the results in Table~\ref{res:setting},
relaxed exclusion outperforms hard exclusion in hierarchical typing
except for Precision when $\theta$ is between $0$ and $0.05$.

\subsubsection{Sample Refinement}
\begin{figure}[h]
\centering
\includegraphics[scale=0.46]{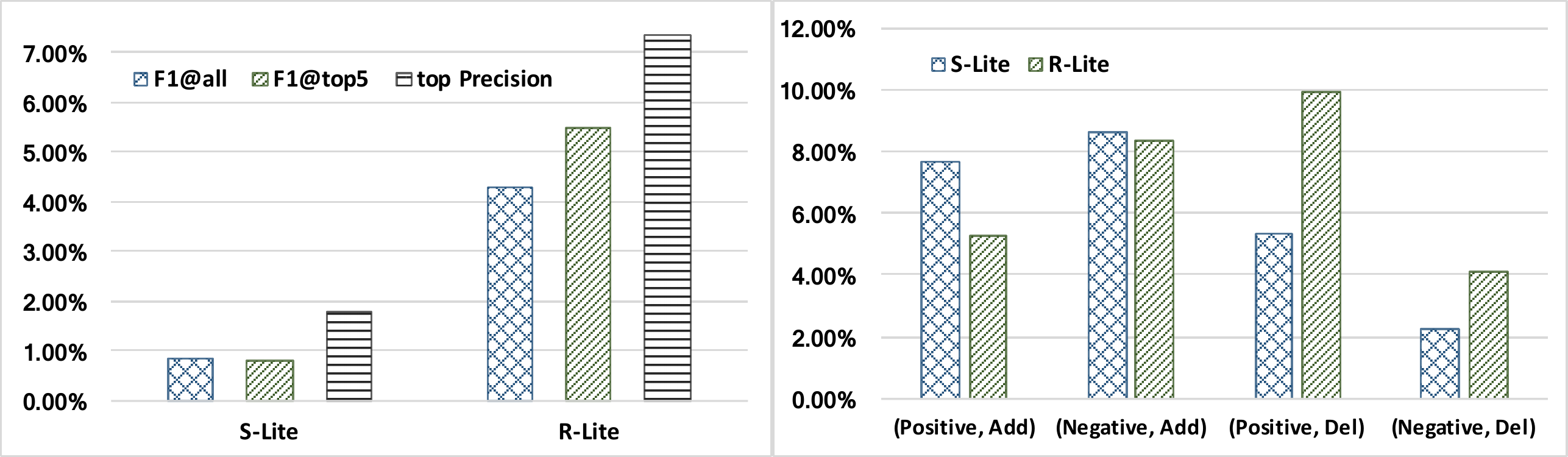}
\caption{\footnotesize
[Left] Performance improvement ($\%$) by sample refinement; [Right] Ratio ($\%$) of added (deleted) positive (negative) particular sample per classifier during sample refinement.
}
\label{res:settings3}
\end{figure}

Fig.~\ref{res:settings3} [Right] shows the ratio of positive and negative particular samples that are deleted and added during sample refinement.
The AttBiRNN classifiers fine tuned by the refined particular samples are compared with those fine tuned by the original particular samples.
The improvements on AvgF1@all, AvgF1@top5 and top Precision, which are based on the average of the three above typing settings, are shown in Fig.~\ref{res:settings3} [Left].
On the one hand, we find sample refinement benefits both S-Lite and R-Lite, as expected.
On the other hand, we find the improvement on S-Lite is limited, while the improvement on R-Lite is quite significant:
F1@all and top Precision, e.g., are improved by around $0.8\%$ and $1.8\%$ respectively on S-Lite, but $4.3\%$ and $7.4\%$ respectively on R-Lite.
This may be due to two factors: 
\textit{(i)} the ground truths of S-Lite are the entities' class and super classes inferred from the KB itself, while the ground truths of R-Lite are manually labeled; 
\textit{(ii)} sample refinement deletes many more noisy positive and negative samples (which are caused by wrong entity classifications of the KB) on R-Lite than on S-Lite, as shown in Fig.~\ref{res:settings3} [Right].

\subsection{Results on Semantic Typing}\label{sec:typing}

Table~\ref{res:typing} displays the overall semantic typing performance of our method and the baselines.
Results for two optimum settings are reported for each method.
The baseline Entity-Lookup retrieves one or several entities using the whole phrase of the literal, and uses their classes and super classes as the types.
Gunaratna \cite{gunaratna2016gleaning} matches the literal's focus term (head word) to an exact class, then an exact entity, and then a class with the highest similarity score.
It stops as soon as some classes or entities are matched. 
We extend its original ``exact entity match" setting with ``relaxed entity match" which means multiple entities are retrieved. 
Property Range Estimation gets the classes and super classes from the entity objects of the property, and calculates the score of each class as the ratio of entity objects that belong to that class.
(H/I, $\kappa$, $\cdot$)@top-P (F1) denotes the setting where the highest Precision (F1 Score) is achieved.
\begin{table}[h!]
\scriptsize{
\centering
\renewcommand{\arraystretch}{1.3}
\begin{tabular}[t]{p{1.7cm}<{\centering}|p{2.3cm}<{\centering}||p{1.2cm}<{\centering}|p{1.2cm}<{\centering}|p{1.2cm}<{\centering}||p{1.2cm}<{\centering}|p{1.2cm}<{\centering}|p{1.2cm}<{\centering}}
\hline
 \multicolumn{2}{c||}{Methods with} & \multicolumn{3}{c||}{S-Lite} & \multicolumn{3}{c}{R-Lite}      \\\cline{3-8}
 \multicolumn{2}{c||}{Their Settings} & Precision & Recall & F1 Score & Precision & Recall & F1 Score \\ \hline
 \multirow{2}{*}{Gunaratna} & exact entity match & $0.3825$ & $0.4038$ & $0.3773$ &$0.4761$ &$0.5528$  &$0.4971$ \\ 
 & relaxed entity match & $0.4176$ & $0.5816$ &$0.4600$&$0.3865$  &$0.6526$ & $0.4469$ \\ \cline{1-8}
 \multirow{2}{*}{Entity-Lookup} & top-$1$ entity & $0.2765$ & $0.2620$ & $0.2623$  &$0.3994$&$0.4407$ &$0.4035$  \\
  & top-$3$ entities & $0.2728$ & $0.3615$ & $0.2962$&$0.3168$ &$0.5201$ &$0.3655$  \\ \cline{1-8}
 Property Range  & (H/I, $\kappa$, $\theta$)@top-P &$0.7563$ & $0.5583$&$0.6210$  &$0.5266$&$0.4015$ 	&$0.4364$ \\
Estimation & (H/I, $\kappa$, $\theta$)@top-F1 &$0.6874$ &$0.7166$ &$0.6773$ &$0.4520$ &$0.5069$ &$0.4632$ \\ \cline{1-8}
  \multirow{2}{*}{AttBiRNN}  & (H/I, $\kappa$, $\theta$)@top-P &$\bm{0.8320}$ &$0.7325$ & $0.7641$ &$\bm{0.7466}$& $0.5819$ 	&$0.6340$ \\
 & (H/I, $\kappa$, $\theta$)@top-F1 &$0.8179$ &$\bm{0.7546}$ &$\bm{0.7708}$ &$0.6759$ & $\bm{0.6451}$&$\bm{0.6386}$ \\ \cline{1-8}
\end{tabular}
\vspace{0.12cm}
\caption{\footnotesize
Overall typing performance of our method and the baselines on S-Lite and R-Lite. 
}\label{res:typing}
}
\end{table}

As we can see, AttBiRNN achieves much higher performance than all three baselines on both S-Lite and R-Lite.
For example, the F1 Score of AttBiRNN is $67.6\%$, $160.2\%$ and $13.8\%$ higher than those of Gunaratna, Entity-Lookup and Property Range Estimation respectively on S-Lite, and $28.5\%$, $58.3\%$ and $37.9\%$ higher respectively on R-Lite.
AttBiRNN also has significantly higher Precision and Recall, even when the setting is adjusted for the highest F1 Score.
This is as expected, 
because our neural network, which learns the semantics (statistical correlation) from both word vector corpus and KB, 
models and utilizes the contextual meaning of the literal and its associated triple, while Gunaratna and Entity-Lookup are mostly based on lexical similarity.    
The performance of Property Range Estimation is limited because the object annotation in DBpedia usually does not follow the property range, especially for those properties in R-Lite.
For example, objects of the property \textit{dbp:office} have $35$ DBO classes, ranging from \textit{dbo:City} and \textit{dbo:Country} to \textit{dbo:Company}.

It is also notable that AttBiRNN and Property Range Estimation perform better on S-Lite than on R-Lite.
The top F1 Score is $20.7\%$ and $46.2\%$ higher respectively, 
while the top Precision is $11.4\%$ and $43.6\%$ higher respectively.
This is because R-Lite is more noisy, with longer literals,
and has more ground truth types on average (cf. Table \ref{res:statistics}), 
while S-Lite has fewer properties, and each property has a large number of entity objects, which significantly benefits Property Range Estimation.
In contrast, the two entity matching based methods, Gunaratna and Entity-Lookup, perform worse on S-Lite than on R-Lite; this is because the construction of S-Lite removes those KB entities from which literals were derived.
Gunaratna outperforms Entity-Lookup as it extracts the head word and matches it to both entities and classes.
Note that the head word is also included in our candidate class extraction with lookup.

\subsection{Results on Entity Matching}\label{sec:entity_annotation}

Table~\ref{res:entity} displays the number of correct matched entities and the Precision of entity matching on R-Lite.
The types are predicted by the fine-tuned AttBiRNN with independent typing and two threshold settings.
We can see that Precision is improved when the retrieved entities that do not belong to any of the predicted types are filtered out.
The improvement is $6.1\%$ and $5.8\%$ when $\theta$ is set to $0.15$ and $0.01$ respectively.
Meanwhile, although the total number of matches may decrease because of the filtering, 
the number of correct matches still increases from $396$ to $404$ ($\theta=0.01$).
This means that Recall is also improved.
\begin{table}[h!]
\scriptsize{
\centering
\renewcommand{\arraystretch}{1.3}
\begin{tabular}{p{2.3cm}<{\centering}|p{1.8cm}<{\centering}|p{2.8cm}<{\centering}|p{2.8cm}<{\centering}}\hline
Metrics&Pure Lookup & Lookup-Type ($\theta = 0.15$) & Lookup-Type ($\theta = 0.01$)   \\\hline
Correct Matches \#     & $396$ &$400$ &$\bm{404}$  \\
Precision     & $0.6781$ &$\bm{0.7194}$ &$0.7176$  \\\hline
\end{tabular}
\vspace{0.18cm}
\caption{\footnotesize
Overall performance of entity matching on R-Lite with and without type constraint.
}\label{res:entity}
}
\end{table}

\section{Related Work}\label{sec:related_work}
Work on KB quality issues can can be divided into KB quality assessment \cite{farber2018linked,zaveri2016quality}, and KB quality improvement/refinement \cite{paulheim2017knowledge}.
The former includes error and anomaly detection methods, such as test-driven and query template based approaches \cite{kontokostas2014test,fleischhacker2014detecting}, with statistical methods \cite{debattista2015quality} and consistency reasoning \cite{paulheim2015serving} also being applied to assess KB quality with different kinds of metric. 
The latter includes \textit{(i)} KB completion, such as entity classification \cite{gangemi2012automatic,paulheim2013type,sleeman2015entity}, relation prediction \cite{krompass2015type} and data typing \cite{dongo2017semantic}; and
\textit{(ii)} KB diagnosis and repair, such as abnormal value detection \cite{fleischhacker2014detecting}, erroneous identity link detection \cite{raad2018detecting} and data mapping (e.g., links to Wikipedia pages) correction \cite{dimou2015assessing}.

KB canonicalization refers to those refinement works that deal with redundant and ambiguous KB components 
as well as poorly expressed knowledge with limited reasoning potential.
Some works in open information extraction (IE) \cite{galarraga2014canonicalizing,vashishth2018cesi,wu2018towards} aim to identify synonymous noun phrases and relation phrases of open KBs which are composed of triple assertions extracted from text without any ontologies.
For example, the recently proposed CESI method \cite{vashishth2018cesi} utilizes both learned KB embeddings and side information like WordNet to find synonyms via clustering. 
Other works analyze synonyms for ontological KBs.
Abedjan et al.\ \cite{abedjan2013synonym} discovered synonymously used predicates for query expansion on DBpedia.
Pujara et al.\ \cite{pujara2013knowledge} identified coreferent entities of NELL with ontological constraints considered. 
\camera{These clustering, embedding, or entity linking based methods in open IE however can not be directly applied or do not work well for our KB literal canonicalization.
The utilization of these techniques will be in our future work.
}

String literals in ontological KBs such as DBpedia often represent poorly expressed knowledge, with semantic types and coreferent entities missed.
As far as we known, canonicalization of such literals has been little studied.
Gunaratna et al.\ \cite{gunaratna2016gleaning} typed the literal by matching its head term to ontology classes and KB entities, 
but the literal context (e.g., the associated subject and property) and semantic meaning of the composition words were not utilized.
Some ideas of entity classification can be borrowed for literal typing but will become ineffective as the context differs.
For example, the baseline Property Range Estimation in our experiments uses the idea of \textit{SDType} \cite{paulheim2013type} --- utilizing 
the statistical distribution of types in the subject position and object position of properties to estimate an entity's type probabilities.
As a literal is associated with only one property, such probabilistic estimation becomes inaccurate (cf. results in Table \ref{res:typing}).

Our literal classification model is in some degree inspired by those natural language understanding and web table annotation works that match external noun phrases to KB types and entities \cite{kartsaklis2018mapping,luo2018cross,chen2019colnet} using neural networks and semantic embeddings for modeling the contextual semantics.
For example, Luo et al.\ \cite{luo2018cross} learned features from the surrounding cells of a target cell to predict its entity association.
However the context in those works is very different, i.e., a simple regular structure of rows/columns with limited (table) metadata. 
In contrast, KBs have a complex irregular structure and rich meta data (the knowledge captured in the KB). 
Differently from these works, 
we developed different methods, 
e.g., candidate class extraction and high quality sampling, 
to learn the network from the KB with its assertions, terminologies and reasoning capability.

\section{Discussion and Outlook}\label{sec:conclusion}

In this paper we present our study on KB literal canonicalization --- an important problem on KB quality that has been little studied.
A new technical framework is proposed with neural network and knowledge-based learning.
It \textit{(i)} extracts candidate classes as well as their positive and negative samples from the KB by lookup and query answering, with their quality improved using an external KB;
\textit{(ii)} trains classifiers that can effectively learn a literal's contextual features with BiRNNs and an attention mechanism;
\textit{(iii)} identifies types and matches entity for canonicalization.
We use a real data set and a synthetic data set, both extracted from DBpedia, for evaluation.
It achieves much higher performance than the baselines that include the state-of-the-art. 
We discuss below some more subjective observations and possible directions for future work.


\subsubsection{Neural Network and Prediction Justification} 
The network architecture aims to learn features from a literal's context.
In our AttBiRNN, a triple is modeled as a word sequence with three size-fixed segments allocated for the subject, object and literal respectively.
The cooccurrence of words and the importance of each word are learned by BiRNNs and the attention mechanism respectively, 
where word position (including whether it is in the subject, property or literal) is significant.
The effectiveness of such a design has been validated in Section \ref{sec:framework}. 
However, the current design does not exploit 
further semantics of the subject, such as its relation to other entities. We believe that this will provide limited indication of the literal's semantic type, but this could be explored using graph embedding methods such as random walks and Graph Convolutional Networks.

We believe that it would be interesting to explore the possible use of the learned attention weights ($\alpha_t$) in justifying the predictions.
For example, considering the literal in triple $\langle$\textit{dbr:Byron\_White}, \textit{dbp:battles}, \textit{``World War II''}$\rangle$ 
and the classifier of type \textit{dbo:MilitaryConflict}, 
\textit{``War''} gets a dominant attention weight of $0.919$,
\textit{``battles''} and \textit{``II''} get attention weights $0.051$ and $0.025$ respectively, 
while the attention weights of other words and the padded empty tokens are all less than $0.0015$.
Similarly, in the triple $\langle$\textit{dbr:Larry\_Bird}, \textit{dbp:statsLeague}, \textit{``NBA"} $\rangle$,
the total attention weights of the subject, property and literal are $0.008$, $0.801$ and $0.191$ respectively w.r.t.\ the classifier of \textit{dbo:Organisation},
but become $0.077$, $0.152$ and $0.771$ w.r.t.\ the classifier of \textit{dbo:BasketballLeague}, where the signal of basketball is focused. 

\subsubsection{Knowledge-based Learning}
We developed some strategies to fully train our neural networks with the supervision of the KB itself.
One strategy is the separated extraction of general samples and particular samples.
It \textit{(i)} eliminates the time consuming pre-training step from a specific task, reducing for example the total typing time per literal of S-Lite from $10.5$ seconds to $2.5$ seconds (training and prediction are run with at most $10$ parallel threads),
and \textit{(ii)} adapts the domain of the classifier toward the target literals through fine tuning, which significantly improves the accuracy as shown in Table~\ref{res:setting}. 
Another strategy that has been evaluated in Section \ref{sec:framework} is sample refinement by validating entity classifications with external knowledge from Wikidata.
However, we believe that this could be further extended with more external KBs, as well as with logical constraints and rules.

\subsubsection{Entity Matching}
We currently search for the corresponding entity of a literal by lexical lookup, and filter out those that are not instances of any of the predicted types.
The extension with prediction does improve the performance in comparison with pure lookup (cf. Section~\ref{sec:entity_annotation}),
but not as significantly as semantic typing, especially on the metric of the number of correct matches.
One reason is that entity matching itself has relatively few ground truths as many literals in R-Lite have no corresponding entities in the KB.
Another reason is that we post-process the entities from lookup instead of directly predicting the correspondence.
This means that those missed by pure lookup are still missed.
In the future we plan to explore direct prediction of the matching entity probably using semantic embedding and graph feature learning.

\section*{Acknowledgments}
The work is supported by the AIDA project (U.K. Government's Defence \& Security Programme in support of the Alan Turing Institute), 
the SIRIUS Centre for Scalable Data Access (Research Council of Norway, project 237889),
the Royal Society,
EPSRC projects DBOnto, $\text{MaSI}^{\text{3}}$ and $\text{ED}^{\text{3}}$. 

%
%
\bibliographystyle{splncs04}
\bibliography{reference}

\end{document}